\documentclass[letterpaper, 10 pt, conference]{ieeeconf}  

\IEEEoverridecommandlockouts                              

\usepackage{cite}
\usepackage{times}
\usepackage{xcolor}
\usepackage{soul}
\usepackage[utf8]{inputenc}
\usepackage{amsmath}
\usepackage{amsfonts}
\usepackage{amssymb}
\usepackage[english]{babel}
\usepackage{algorithm}
\usepackage{algpseudocode}
\usepackage{multirow}
\usepackage{graphics} 
\usepackage{epsfig} 

\usepackage{graphicx}
\usepackage{caption}
\usepackage{subcaption}

\title{\LARGE \bf
Learning Coordinated Tasks using Reinforcement Learning in Humanoids
}
\author{Phaniteja S$^{1*}$, Parijat Dewangan$^{1*}$, Pooja Guhan$^1$, K Madhava Krishna$^1$, Abhishek Sarkar$^1$
\thanks{*These authors have contributed equally.}
\thanks{$^{1}$All the authors are with Robotics Research Center, International Institute of Information Technology, Hyderabad, India
        {\tt\small  \{phaniteja.sp, parijat10, poojaguhan97\}@gmail.com, \{mkrishna, abhishek.sarkar\}@iiit.ac.in}}%
}

\begin{document}

\maketitle
\thispagestyle{empty}
\pagestyle{empty}

\begin{abstract}
With the advent of artificial intelligence and machine learning, humanoid robots are made to learn a variety of skills which humans possess. One of fundamental skills which humans use in day-to-day activities is performing tasks with coordination between both the hands. In case of humanoids, learning such skills require optimal motion planning which includes avoiding collisions with the surroundings.
In this paper, we propose a framework to learn coordinated tasks in cluttered environments based on DiGrad - A multi-task reinforcement learning algorithm for continuous action-spaces. Further, we propose an algorithm to smooth the joint space trajectories obtained by the proposed framework in order to reduce the noise instilled during training.
The proposed framework was tested on a 27 degrees of freedom (DoF) humanoid with articulated torso for performing coordinated object-reaching task with both the hands in four different environments with varying levels of difficulty. It is observed that the humanoid is able to plan collision free trajectory in real-time. Simulation results also reveal the usefulness of the articulated torso for performing tasks which require coordination between both the arms.
\end{abstract}

\section{INTRODUCTION}

Humanoid robots has constantly fascinated the research society since the past few decades. Although numerous humanoid robots have been developed till date, motion planning for such a complex robotic system still poses many challenges. \cite{hauser2005non,dalibard2009whole,teja2016optimal} used the sampling based algorithms like PRM \cite{kavraki1996probabilistic}, RRT-connect \cite{kuffner2000rrt}, RRT*-connect\cite{klemm2015rrt} etc. for motion planning of humanoid robots in different environmental settings.  \cite{el2013optimal} proposed an optimal motion planner by posing it as optimal control problem and combining it with RRT. Although these motion planners guarantee the optimal path, they are computationally expensive and thus cannot search for the optimal path in real-time. Besides, all the mentioned works in path planning considered only static obstacles. \cite{budiharto2013framework} proposed a stereo vision based approach for dynamic obstacles avoidance in humanoid robots with predefined motions for manoeuvre.
In the famous humanoid robot challenge conducted by DARPA, humanoids were required to perform a variety of manoeuvres and coordinated tasks which shows the need for real time controllers and motion planning in complex 3D environments. 
\begin{figure}
\centering
\includegraphics[scale =0.385]{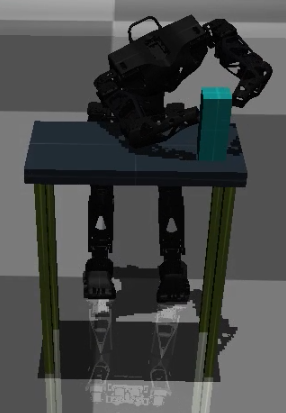}
\includegraphics[width=0.5\columnwidth]{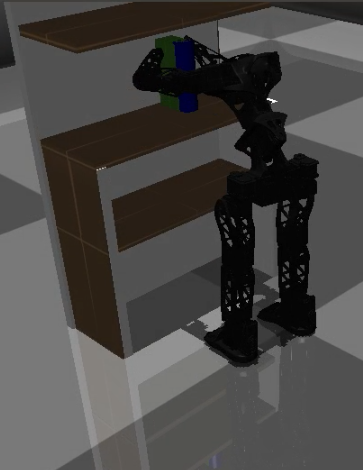}
\caption{Humanoid with articulated torso grabbing an object}
\label{intro_pic}
\end{figure}

Coordinated tasks in humanoid robots are highly dependant on the structure of the robot. In many humanoids like ASIMO \cite{sakagami2002intelligent}, KHR \cite{kim2005system} and HUBO\cite{park2007mechanical} series robots of KAIST, the HRP\cite{kaneko2009cybernetic} series robots of AIST, iCub\cite{hinojosa2006performance} the entire upper body is built using a box shaped torso with very limited DoF. Hence they are accompanied by arms with 6-7 DoF to perform complex manipulation tasks. Another set of robots which are inspired by human, like Acroban \cite{lapeyre2013physical}, Poppy\cite{poppy} use articulated torso and arms with lesser DoF. Although the presence of articulated torso makes the robot motion close to human, it also raises several problems. Besides increased complexity in terms of design (control, power), the torso joints are usually exposed to higher torques as they balance the entire upper body. The requirement of stronger motor makes the torso heavy, thus limiting robot's physical size as well.  
On the positive side, the presence of articulated torso increases the configuration space of the robot. In our work, we show how the articulated torso helps in performing coordinated tasks which involves motion planning of the whole upper body.

Modelling dynamics in high-dimensional robots like humanoids is difficult. Thus, motion planning in these robots may not be real-time due to complex computations involved in solving dynamics. With the advancement of Machine Learning (ML) and Artificial Intelligence (AI) in the recent times, several frameworks were proposed for accomplishing complex control tasks in 3D environments.  \cite{morimoto2004simple} used reinforcement learning (RL) \cite{sutton} to teach a biped robot to walk. \cite{peters2003reinforcement} proposed a natural actor-critic algorithm for learning motor control in humanoid robots. \cite{mombaur2010human} proposed an inverse optimal control approach to learn humanoid location from humans. All these works show the robustness of RL for solving high-dimensional motion planning in humanoid robots. Recent advances in Deep RL like Deterministic Policy Gradients(DPG)\cite{silver2014deterministic}, \cite{lillicrap2015continuous} introduced Deep RL for continuous control tasks which further extended the scope of RL in robotics. 

In this paper, we contribute a novel architecture DiGrad as explained in \cite{digrad}, based on differential policy gradients to learn coordinated tasks in complex 3D environments. This is also one such application of Deep RL for solving complex problem of motion planning of articulated torso in constrained workspace. We show repeatable goal reaching behaviour in diverse environments in presence of obstacles of various shapes and sizes, thereby indicating the suitability of DiGrad to the complex problem of planning in high dimensional space. Also, unlike sampling based planners, we bypass the need for accurate and robust dynamics modelling of the system. 

\begin{figure}
\centering
\includegraphics[width=\columnwidth]{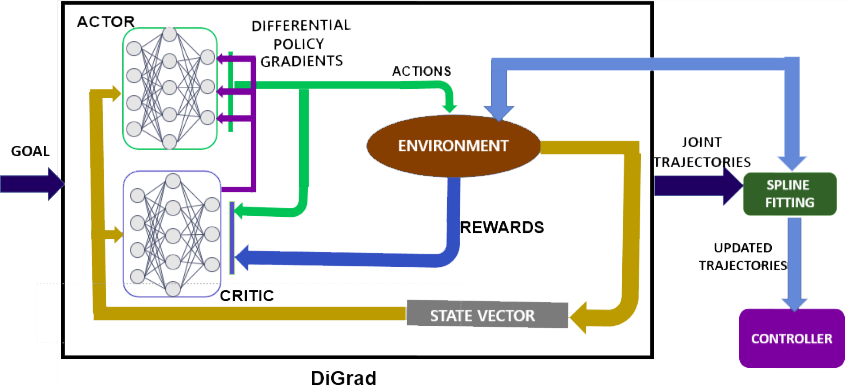}
\caption{Pipeline for motion planning}
\label{digrad}
\end{figure}
Although the solutions obtained using Deep RL algorithms are robust, they suffer from noisy outputs which should be addressed  especially in bipedal robots as they are prone to lose balance easily. Hence, we propose a novel binary search based spline fitting algorithm for smoothing of joint trajectories as obtained from RL based solvers. This approach ensures that the final configurations are kept same. This approach also takes into account collision avoidance and joint limit constraints while solving for smooth joint trajectories. The algorithm run-time complexity is $O(log(n))$ which is much lesser as compared to complex optimization methods used to solve this issue. The main contribution of this paper is the entire pipeline for motion planning in complex 3D environments as shown in Fig. \ref{digrad}, starting from learning coordinated tasks in cluttered environments using Deep RL to obtaining smooth joint trajectories. The final obtained trajectories are stable and ensure collision avoidance with the surroundings. 

The rest of the paper is organised as follows. Section II talks about the role of articulated torso in coordinated tasks followed by the description of our robot. Section III explains DiGrad and shows how this can be used to learn complex coordinated tasks by suggesting the modifications to state vectors and reward function. Section IV discusses the proposed methodology for smoothing of joint trajectories followed by section V showing the training and simulations results. Finally, conclusions are given in Section VI.

\section{HUMANOIDS WITH ARTICULATED TORSO FOR COORDINATED TASKS}
Humanoid robots are large electro mechanical systems which hold the capability to perform complex tasks in complex environments. Although humanoid robots are in existence since a long time, making the robot to perform coordinated tasks is still one of the challenging areas of research. Several designs have been proposed for human-like robots. The articulated torso design is one of most interesting design as it resembles the human spine. In this section, we discuss the importance of articulated torso for coordinated tasks following by the description of our humanoid robot.
\subsection{Role of articulated torso in coordination}
The ability to efficiently move and perform challenging tasks in complex environments is a fundamental requirement for humanoid robots and the torso of the robot plays a crucial role in accomplishing such tasks. In case of traditional humanoids, the box shaped torso limits their workspace and hence their ability to do complex reachability or manoeuvrability tasks. Such a torso contains less DoFs in the upper body and doesn't offer enough flexibility to perform complex coordinated tasks.
In order to overcome these limitations, roboticists started taking inspiration from nature and developed humanoid robots with an articulated torso. With the additional DoF provided by the articulated torso, a larger range of body configurations are available for the robot that in turn enhances flexibility, stability and manoeuvrability. Specially while performing coordinated tasks, an articulated torso can provide more natural solutions as it has a large configuration space. Hence in all of our experiments, we have considered a humanoid robot with articulated torso, whose details are discussed below.
\subsection{Humanoid Robot Model}
Our humanoid robot\cite{goel2016design} shown in Fig. \ref{intro_pic} is a modified version of the Poppy robot \cite{poppy}. The design has been modified so that the robot can handle a task with more efficiency and power. The robot has 27 DOF. The height is approximately 84 cm which is roughly half of the average human height. The entire robot is 3D printed with PLA material which makes it sturdy without adding up too much weight to the total structure keeping it lightweight. The entire structure of the humanoid robot can be divided into different sections - legs (6 DoF each), hands (4 DoF each), torso (5 DoF) and head (2 DoF). The torso with 5 joints emulates a simpler vertebral column that acts as a joining link between the shoulders and pelvis.
The presence of such a flexible spine and high DoF leads to the enhancement of human-like movements of the robot and increases the range of movement of the robot's torso and also hence increases its effective work space. Therefore planning using sampling based solvers become computationally expensive and in many cases very slow. Learning based approaches provide a very good alternative to learn such complex behaviours in humanoid robots.  

\section{DEEP RL FOR COORDINATED TASKS IN CLUTTERED ENVIRONMENTS}

RL is a very useful technique to teach complex tasks to humanoid robots. 
In any standard RL environment, there is an agent and an environment. The agent in state $s$ performs an action $a$ on the environment and the environment gives out the updated state $s'$ and the reward $r$ back to the agent. The main goal of the RL is to maximize the cumulative reward in an episode. An episode is described as a sequence of states, actions and rewards, which ends with terminal state. 
Development of elegant Deep RL algorithms \cite{ddpg,dpg} in recent times provided a framework for learning continuous control of manipulators\cite{gu2017deep}. Inspired by this \cite{phaniteja2017deep} applied the Deep RL framework to reachability tasks in humanoid robots. 

DiGrad is a simultaneous multi-task learning algorithm proposed for solving continuous control problems in complex robotic systems. It provides us a baseline for solving reachability tasks in multi-armed manipulators.
In this section, we explain how DiGrad can be modelled for learning coordinated tasks in humanoid robots. Further we explain how this can be easily extended to cluttered environments to learn collision avoidance.
\subsection{DiGrad}
DiGrad is an actor-critic based reinforcement learning algorithm for performing control tasks in continuous action spaces. The algorithm uses differential policy gradient update for actor instead of the standard policy gradient update suggested by DPG. This makes the algorithm robust and increases the convergence rate. Consider $k$ tasks in a standard reinforcement learning environment that are needed to be learnt together and have the same state space. Let the critic network $Q$ be parametrized by $\theta^{Q}$ and actor network $\mu$ by $\theta^{\mu}$. The critic network $Q$ outputs multiple action-values each corresponding to a task. DiGrad aims to learn the compound policy $\mu$ by splitting it into multiple sub-policies $\mu_i$ and taking a combined update using these sub-policies. Suppose $Q_i$ is the action-value function corresponding to $i^{th}$ task obtained by performing action $a_i$ in state $s$, then the policy gradient according to DiGrad is:
\begin{equation}
\label{policy_gradient}
\nabla_{\theta^\mu} J \approx \sum_{i=1}^{k}\mathbb{E}[\nabla_{a_i}{Q_i(s,a_i|\theta^Q)}\nabla_{\theta^\mu}\mu_i(s|\theta^{\mu})]
\end{equation}
where $a_i = \mu_i(s|\theta^{\mu})$ and $\mathbb{E}$ refers to expected value. Now the update on actor with learning rate $\eta$ is given by:
\begin{equation}\label{actorupdate}
\theta^{\mu} = \theta^{\mu} + \eta(\sum_{i=1}^{k}\mathbb{E}[\nabla_{a_i}{Q_i(s,a_i|\theta^Q)}\nabla_{\theta^\mu}\mu_i(s|\theta^{\mu})])
\end{equation}
There are some cases where a set of actions affect more than one tasks, we call this set of actions as shared actions $a_s$. Let's consider the case of the 27 DoF humanoid as shown in Fig. 1 . As mentioned in the previous section, it has a multi-chain architecture for the upper body  where the 5 DoF spine is shared among the two chains. Thus, the spine contributes to the reachability tasks of both the hands. Therefore, we can say that the spine acts as a shared action for tasks that require coordination of both the hands. In such cases where a set of actions $a_s$ are shared between all the tasks, we use the policy gradient update as mentioned in Eq. \eqref{hu}.
\begin{equation}\label{hu}
\hspace{-0.5cm} 
\begin{split} 
\nabla_{\theta^\mu} J \approx  \sum_{i=1}^{k}\mathbb{E}
[\nabla_{a^d_i}Q_i(s,a_i|\theta^Q)\nabla_{\theta^\mu}\mu^d_i(s|\theta^\mu)] \\ + 
\frac{1}{k}\mathbb{E}[\sum_{i=1}^{k}\nabla_{a_s}Q_i(s,a_i|\theta^Q)\nabla_{\theta^\mu}\mu_s(s|\theta^\mu)]
\end{split}
\end{equation}
where $a_i^d = \{a_i - a_s\}$ and $\mu^d_i = \{\mu_i - \mu_s\}$.

Let the reward for the $i^{th}$ task be denoted as $r_i(s,a_i)$ and $\gamma$ be the discount factor. The agent performs an action $a_i$ on the environment in state $s$ and moves to new state which we denote by $s'$. In order to stabilize the learning, target networks $\theta^{Q'}, \theta^{\mu'}$ are used for both critic and actor as suggested by DDPG and the critic network is updated by minimising the following loss function:
\begin{equation}\label{loss}
L(\theta^{Q}) = \sum_{i=1}^{n} (Q_i(s,a_{i}|\theta^{Q}) - y_{i})^2
\end{equation}
where $y_i$ is the target given by:
\begin{equation*}\label{target}
y_{i} = r_i(s, a_i) + \gamma Q(s',{\mu}(s'|\theta^{\mu'})|\theta^{Q'})
\end{equation*}
Lastly, the target networks are updated by using $\tau << 1$ as follows:
\begin{equation}\label{targetupdate}
\begin{split}
\theta^{Q'} = \tau \theta^{Q'} + (1-\tau)\theta^Q\\
\theta^{\mu'} = \tau \theta^{\mu'} + (1-\tau)\theta^\mu\
\end{split}
\end{equation}
In order to apply DiGrad for our humanoid robot, we have chosen the preliminary state vector $s$ containing all joint angles $q_i$, the present Cartesian space coordinates of the end effectors (hands) and the Cartesian space coordinates of the goal positions, for learning coordinated tasks. The compound action vector consists of the joint angular velocities $\dot q_i$ needed to reach the goals. The reward function and the complete state vector modelling is explained in the next subsection.   
\subsection{Extending DiGrad for Cluttered Environments}
In order to apply DiGrad for cluttered environments, we have to take into account collision avoidance along with goal reachability. Further, we should ensure that both the hands reach the goal positions simultaneously. Hence the reward function modelling becomes a crucial part in the learning process. Based on all this criteria we modelled the reward function as follows:
\begin{equation}\label{reward}
 r_i = -\alpha dist_i + 
\begin{cases}
-n_1 \ if(cols)\\
-n_2 \ if(instb)\\
+m_1 \ if(gb_i)\\
+m_2 \ if(goal_i)\\
\end{cases}
\end{equation}
where $cols$, $instb$, $gb_i$, $goal_i$ are flags referring to collision, instability, goal boundary and goal respectively and $n_1,n_2,m_1,m_2$ are positive constants. 

In order to ensure that both the hands reach the end goal position simultaneously as is needed for the coordinated tasks, we have taken the above reward function for each arm of the humanoid and a very large positive reward $\kappa$ is given when both the hands reach the goal simultaneously which encourages the robot to learn the coordinated goal reachability. Hence,
\begin{equation}
\begin{split}
\quad r_1 = \kappa, r_2 = \kappa \quad if(goal_1 \quad and \quad goal_2)
\end{split}
\end{equation}

In the case of cluttered environments, we can use different types of state vectors depending upon the type of the obstacles. In this paper, we explore two different types of environment.  In order to learn the entire workspace with obstacles of varying position and size, we include the position and size of the obstacles into the state vector $s$. Note that, the number of obstacles is kept constant. During training, the position and size of the obstacle are randomly sampled in each episode and the robot is trained to reach the goal avoiding collisions with the obstacle along with maintaining stability. This kind of setting allows the robot to be robust and adaptable to unseen environments as well. In another setting, we have considered the environment to be constant throughout where the obstacles are large static objects like shelf. In this setting, the obstacle data need not be included in the state vector as it is constant. In both these settings, the goal position changes randomly in the workspace in each episode. 

\subsection{Training and Environment}
MuJoCo\cite{todorov2012mujoco} is a dynamic simulation environment which can do efficient rigid body simulations with contacts.
We used MuJoCo as an environment for training in all settings. MuJoCo provides us with accurate collision and position data, which we use to model our state and reward function. The training is carried out by executing actions provided by the actor in MuJoCo and observing the obtained states and rewards. The entire training loop is shown in Algorithm 1. Note that, the episode is terminated whenever the robot loses its balance or when the goals are reached.  
\begin{algorithm}
\caption{DiGrad Training}\label{ddpg} 
\begin{algorithmic}[1]
\Statex Randomly initialize actor $\theta^\mu$ and critic $\theta^Q$ networks
\Statex Initialize target network as $\theta^{Q'} \gets \theta^Q$ and $\theta^{\mu'} \gets \theta^\mu$
\Statex \textbf{for} ep = 1 to MaxEp \textbf{do}
\Statex \quad Randomly sample goal object position.
\Statex \quad Randomly sample obstacles positions (if needed).
\Statex \quad Set goals to each arm and get state $s$.
\Statex \quad Reset the flags $cols,instb,gb_1,gb_2,goal_1,goal_2$.
\Statex \quad Check for collision and stability.
\Statex \quad \textbf{for} st = 1 to MaxStep \textbf{do}
\Statex \qquad Get action $a$ = actor($s$) + $N$ (noise)
\Statex \qquad Execute action $a$  and set the appropriate flags.
\Statex \qquad Get rewards $\vec{r}$ using Eq. \eqref{reward} and new state $s'$
\Statex \qquad Store $s,a,s',\vec{r}$ in replay buffer R.
\Statex \qquad Update networks using Eq. \eqref{actorupdate}, \eqref{hu}, \eqref{loss} and \eqref{targetupdate}.
\Statex \qquad \textbf{if} ($goal_1\ and \ goal_2$) or $instb$ \textbf{then}
\Statex \qquad\quad break;
\Statex \qquad \textbf{end if}
\Statex \quad \textbf{end for}
\Statex \textbf{end for}
\end{algorithmic}
\end{algorithm}

In order to explore the workspace,  an exploration noise $N$ needs to be added to the action during training. In our case,  we have taken a decaying random normal noise as $N$. We use experience replay which addresses the issue of data being dependent as most of the optimization algorithms need samples which are sampled from identically independent distributions (i.i.ds). Hence, we use a replay buffer R which stores the data of every step. For training, we randomly sample from the replay buffer which ensures that samples drawn are from identically independent distributions. Once the training is complete, the solutions are tested without added noise. The solutions obtained are collision free and stable but have oscillations due to the noise added during training. In the next section we propose an algorithm to reduce these oscillations.

\section{Smoothing of joint trajectories}
RL based solvers for continuous space are usually trained with added noise in order to increase the exploration of the workspace. Thus, the solutions by given RL framework in continuous space are usually noisy. In the case of humanoids which is highly susceptible to lose balance, these noisy solutions might lead to unwanted jerks which makes the robot less stable and also result in discontinuous motor profiles which increases the torque requirements of the motor significantly. Thus, the joint space trajectory found in Section III should be further filtered and smoothed, in order to remove the noise and also reduce the torque requirements. Note that, we also need to ensure that the smoothing doesn't lead to collision. In this section, we propose an algorithm to smooth the joint trajectories provided by the framework in Section III, such that the smoothing doesn't lead to instability or collision.

\subsection{Spline fitting}

Let's say, we have the joint space trajectories for all the joints, which we call as knot points. As mentioned in Section III, these joint space trajectories are collision free and ensure stability. We need to fit a spline function $f$ for all joints such that it is continuous and  has continuous derivatives at its knot points. 
In our case, we use cubic splines\cite{de1978practical} which are piecewise cubic functions that are continuous and have continuous first and second derivatives. Let $J^k$ be a set of joint space trajectories for some joint k as given by DiGrad. The cubic smoothing spline estimates function $f^k$ which minimizes  
\begin{equation}\label{cost_function}
p\sum_{t=0}^{T}w_t(J^k_t - f^k(t)) + (1-p)\int (\frac{d^2f^k}{d t^2})^2 dt
\end{equation}
where $J^k_t$ is joint angle of joint $k$ at time $t$, $w_t$ is the specified weight for time t and p is the smoothing parameter, which controls the trade-off between fidelity to the data and roughness of the function $f^k$. Thus, p $\rightarrow$ 0 is infinite smoothing where the estimate converges to linear least squares estimates and p $\rightarrow$ 1 is no smoothing where the estimate converges to interpolating spline. Also, the weight for t=0 and t=T are given a higher value in order to ensure that the initial and final joint angles are same, thus completing the task.

\begin{algorithm}
\caption{Checks if smoothing parameter provided is within solution space}
\begin{algorithmic}
\State \textbf{INPUT:} J, k, p \Comment{Trajectories  of all joints, joint index and smoothing parameter }
\State
\Procedure{Evaluate}{}
\State $J^k$ = $J[k]$ \Comment{Trajectory of joint k}
\State $J^k_{new}$ = \textbf{CubicSpline}($J^k$, p)
\State Update $J$ with $J^k_{new}$
\If{J is collision free} \Comment{Check constraints}
\State return true;
\Else{}
\State return false;
\EndIf
\EndProcedure
\end{algorithmic}
\end{algorithm}
\begin{algorithm}
\caption{Binary Search for optimal smoothing parameter}
\begin{algorithmic}
\State \textbf{INPUT:} J, joint \Comment{Trajectories  of all joints and current joint }
\State
\Procedure{BinarySearch}{}
\State lower = 0, upper = 1 \Comment{Initialise lower \& upper limit}
\While{lower \textless  upper}
\If{upper - lower \textless  precision}
\State break
\EndIf
\State mid = (lower + upper) / 2
\If {\textbf{Evaluate}$(J,joint, mid)$ is true}
\State upper = mid
\Else
\State lower = mid
\EndIf
\EndWhile
\State \textbf{return} $end$
\EndProcedure
\end{algorithmic}
\end{algorithm}
\begin{algorithm}
\caption{Spline fits the provided joint trajectories}\label{alg:sf}
\begin{algorithmic}
\State \textbf{INPUT:} J \Comment{Trajectories of all joints}
\State
\Procedure{SplineFit}{}
\For {each joint starting from base}
\State $J^k$ = $J[joint]$ \Comment{Trajectory of current joint}
\State {$p_{opt}$} = \textbf{BinarySearch}($J$, $joint$)
\State {$J^k_{new}$}= \textbf{CubicSpline}($J^k$, $p_{opt}$)
\State Update $J$ with $J^k_{new}$
\EndFor
\State return $J$
\EndProcedure
\end{algorithmic}
\end{algorithm}

\subsection{Search for the optimal smoothing factor}
For each joint trajectory, we need to fit a spline with the most optimal smoothing parameter. The smoothing factor $p$ can vary from 0 to 1. The optimal smoothing parameter for spline is the one which maximises the smoothing along with constraint as collision avoidance in the environments.  Let's say the optimal smoothing parameter is $p_{opt}$. From objective function of the cubic spline Eq. \eqref{cost_function}, higher the smoothing parameter $p$, more is the closeness to the original joint space trajectory and thus lesser chances of collision. Thus, we can say that $p \in [p_{opt},1]$ is the solution space where the constraints are satisfied, whereas in $p \in [0,p_{opt})$ the constraints are not satisfied. The solution space is defined as the range of $p$, for which the smoothen trajectories does not lead to collision or instability. The optimal smoothing parameter $p_{opt}$ can be calculated with a modified binary search algorithm on the parameter $p$ as given in Algorithm 3. In this procedure, we start by initializing the upper and lower bounds of $p$. The search begins by evaluating the middle value, if it lies within the solution space or not. If the middle value is in the solution space, the search continues in the left half, else, the search continues in the right half. In this way, we keep on discarding one of the halves until we reach the optimal smoothing parameter.

\begin{figure*}[ht!]
\label{env_tsh}
\centering
\includegraphics[width=2\columnwidth]{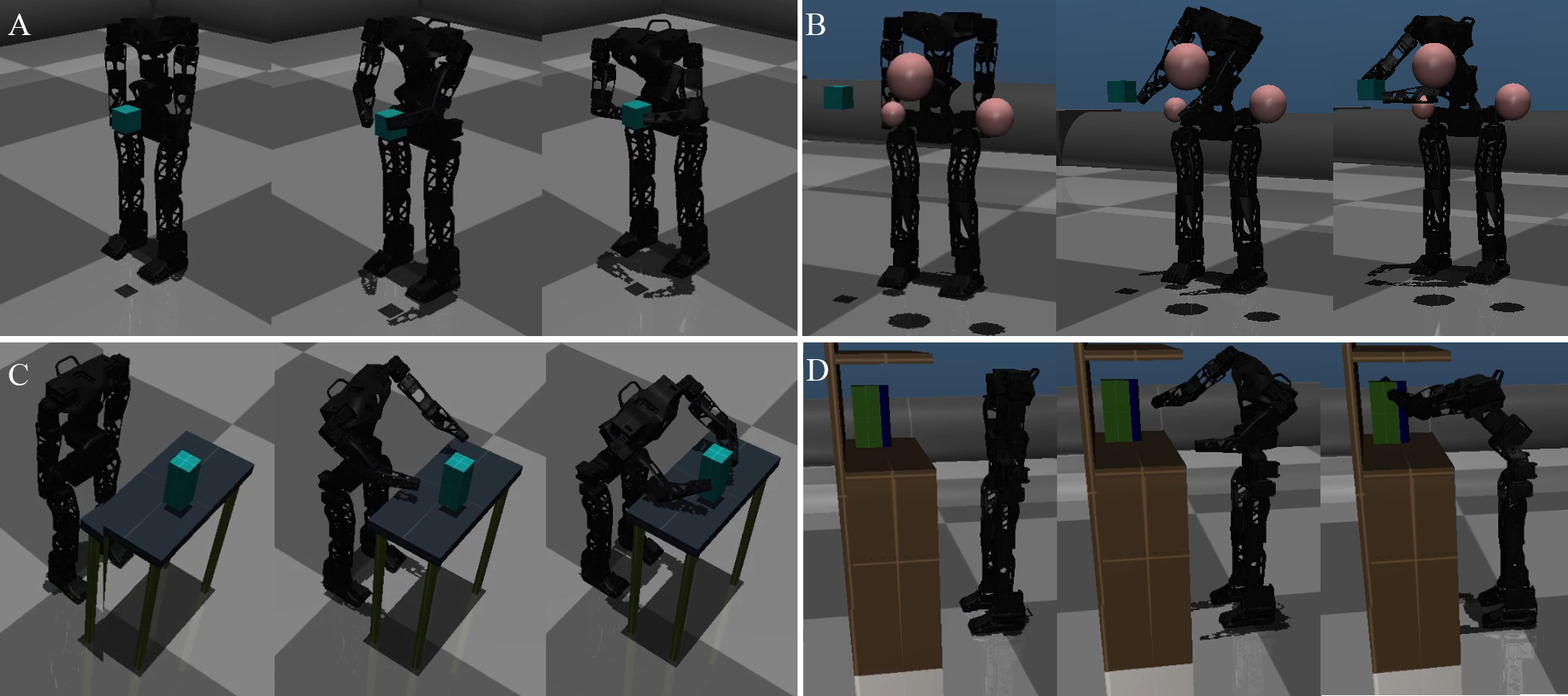}
\caption{ Humanoid robot performing reachability tasks in different environments (A) No obstacles (B) Random Obstacles around the target position (C) Grabbing block from a table (D) Reaching out to a book located in a shelf }
\end{figure*}
\begin{figure*}[t]
        \begin{subfigure}[h]{0.24\textwidth}
			\includegraphics[width=\textwidth]{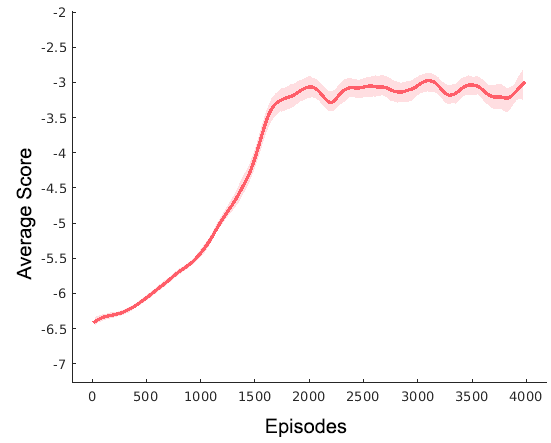}
			\caption{No obstacles}
			\end{subfigure}
        \begin{subfigure}[h]{0.24\textwidth}
				\includegraphics[width=\textwidth]{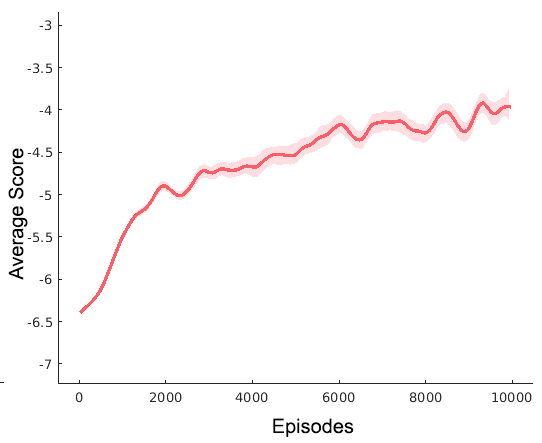}
                \caption{3 sphere-shaped obstacles}
        \end{subfigure}
                \begin{subfigure}[h]{0.24\textwidth}
			\includegraphics[width=\textwidth]{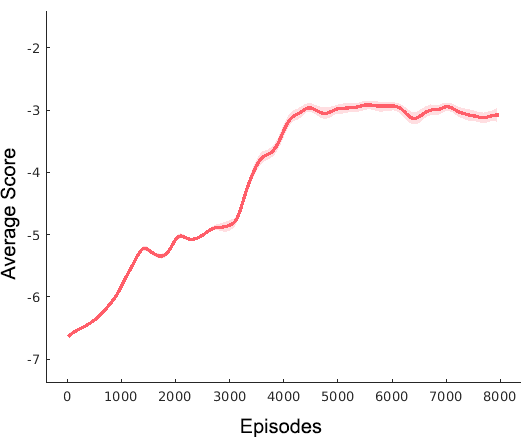}
			\caption{Table}
			\end{subfigure}
        \begin{subfigure}[h]{0.24\textwidth}
				\includegraphics[width=\textwidth]{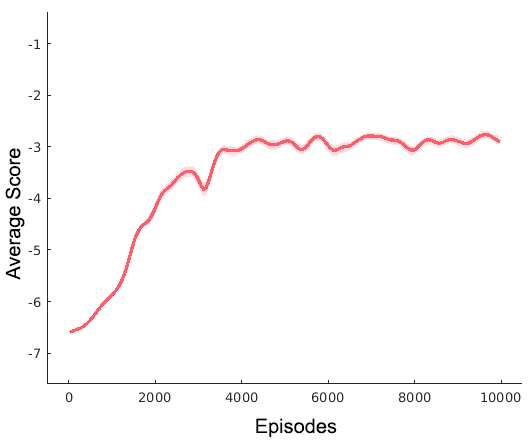}
                \caption{Shelf}
        \end{subfigure}
        \caption{Figure shows the performance curves of the training in different environment settings. In all the graphs, the x-axis shows the number of training episodes. The bold line represents the mean value of the score and the coloured area around it shows the standard deviation of the score.}
\end{figure*}
\begin{figure*}[ht!]
\label{traj_1}
\centering
\includegraphics[width=2\columnwidth]{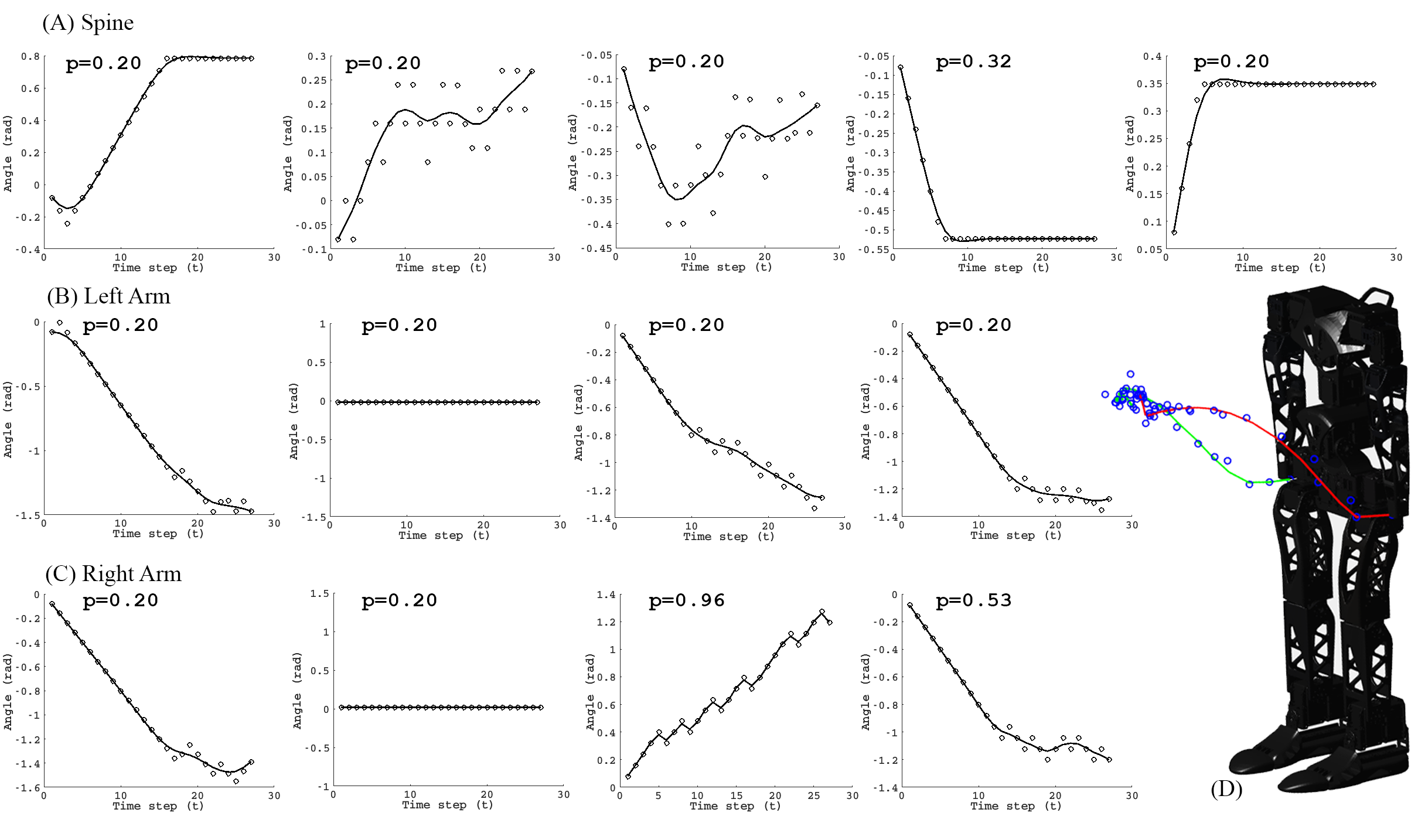}
\caption{(A)-(C) show the joint space trajectories for spine, left arm and right arm respectively. (D) shows the knot points and the final smoothed out trajectories of both the end effectors (hands) of the humanoid robot. In all the graphs the knot points are shown using circular dots and the smoothed trajectories are shown using a solid line.}
\end{figure*}

Algorithm 4 explains the procedure for computing smooth joint trajectories using \textbf{Evaluate} and \textbf{BinarySearch}. \textbf{SplineFit} function iterates over all the joints starting from base joint and finds the $p_{opt}$ for each joint. Note that, $J$ contains the trajectories of all the joints and is updated after every iteration with the new trajectory of the current joint. Finally at the end of the search, $J$ contains the updated trajectories of all the joints.
\textbf{CubicSpline} fits the provided knot points minimising the cost function as stated in Eq. \eqref{cost_function}. \textbf{Evaluate} fits the trajectory of the provided joint index according to the provided smoothing parameter $p$ and checks whether the new trajectory is within the constraints provided, that are collision avoidance and stability. \textbf{BinarySearch} finds the optimal smoothing parameter for the given joint. The time complexity for finding $p_{opt}$ for all the joints is $n$ * O(-$log_2$($precision$)), where $n$ is number of joints and $precision$ is set as $10^{-6}$.  

\section{RESULTS AND SIMULATIONS}
\subsection{Environments}
The proposed framework was tested for reachability tasks where the robot learns to grab an object using both the hands, taking into consideration collision avoidance and stability criteria. The framework was tested in different environments with varying difficulty levels to test the  robustness of the framework. The following are the different environments (Fig 3) where the robot learns to perform coordinated tasks keeping in consideration the surroundings.
\subsubsection{A simple 3D environment with no obstacles} This is the easiest of all the environment where the robot is trained to grab a cube with  both the hands. The cube is sampled at random position in the configuration space of the robot in each episode. Thus, the robot learns to grab an object anywhere in its configuration space. 

\subsubsection{3D Environment with Random Obstacles} This setting is similar to the above one except for the 3 sphere shaped obstacles which are included in the environment while the robot tries to grab the cube. The positions and sizes of these spheres are sampled randomly at each episode. This randomization of the obstacles adds robustness and adaptability to the obtained solution. 
\subsubsection{Table} In this setting, the environment consists of a table and the objects are sampled at random positions on the top of the table every episode. The robot learns to grab a cube while trying to avoid collisions with the table. Unlike the previous environment, the obstacle i.e. table details are not included in the state vector as it is constant. Learning collision avoidance is more complex than the previous environments as the obstacle is large in size and within very close vicinity of the goal object.   
\subsubsection{Shelf} This setting is similar to the Table environment but even more complex since the configuration space of robot is highly restricted due to large number of possible collisions. The multiple racks increase the difficulty level of learning as all the body parts of the humanoid are collision prone in this case. Here, the robot tries to grab a book in the shelf whose position changes in each episode.
\subsection{Training}
In all the experiments, the agents were implemented in a TensorFlow codebase. Both the actor and critic networks consist of two fully connected hidden layers. Hidden layers consists of 700 and 400 hidden units with CReLu activation and a drop-out of 0.8. Batch normalization is used in actor network at all layers but only in the first hidden layer in critic network. In critic network, L2 regularization of 0.01 is also used. The learning rate for both actor and critic is taken as $10^{-4}$. The replay buffer size is set as 45000 for all the settings. The training for each environment is run for at least 0.6 million steps and at most 1.5 million steps depending on the setting. For the environment without any obstacles, the training was run for 4000 episodes while for the remaining environments it was run for 10000 episodes. In both the cases, each episode was run for 150 steps. For all the experiments, we define the quantities error and score as follows:
\begin{equation}
\begin{split}
error_i = ||G_i-E_i||,\ i=1,2 \\
score = -log(error_1 + error_2)
\end{split}
\end{equation}
where $G_i$ and $E_i$ are the coordinates of the goal end end effector of the $i^{th}$ hand. The training performance curves for all the environments are shown in Fig. 4. From these curves, we can say that the learning saturated at 4000 episodes for all the environments and the robot was able to reach the goal avoiding the obstacle in their respective environment. 

\subsection{Spline Fitting Results}
The joint angle trajectories obtained from the RL framework are smoothed using the proposed binary search based algorithm as given in Algorithm 3. The plots of the spline fitted joint trajectories with their knot points as obtained by RL and their respective smoothing parameter $p$ for one of the cases of Table environment are shown in Fig. 3(C).
In Fig. 5(D), the end effector trajectories obtained after applying smoothing to the joint trajectories is shown. We can see that the improved hand trajectories have much lesser oscillations as compared to the initial trajectories. It can be observed from the Fig. 5(A-C) that the final smoothed trajectories do not pass through all the knot points but maintains the final configuration which is of high necessity for completing the task. Note that, the smoothing takes into consideration collision avoidance in its environment, while optimizing. 

\subsection{Simulation Results}
Fig. 3 shows the test results where the robot is trying to grab an object with both the hands in all the environments explained above. Each of these environments differ in the way collision avoidance is learnt by the actor. In case of dynamic obstacle environment (Fig. 3B), the obstacle details are included in the state vector. Thus, the DiGrad network computes the trajectories keeping in consideration the obstacle position and size. In the case of Table and Shelf environments, the network memorize the entire workspace as the obstacles are constant. The trajectory generation in these cases needs to be precise as the obstacles are closely clustered near the goal object. Fig. 5(D) shows the  hand trajectory for grabbing the goal object in Table environment. We can see that the left and right trajectories are almost symmetric reaching the goal simultaneously.  It can be observed from Figs. 1 and 3 that the articulated torso plays an important role in helping both hands reach their goal position simultaneously as well as to maintain stability throughout the process. 

\section{Conclusion} 
In this work, we showcased a novel framework for learning and performing coordinated tasks in clustered environments in humanoid robots. DiGrad learns the optimal joint trajectories required to accomplish the given task. These trajectories are usually accompanied by the noise instilled during the training process which reduces the stability of the humanoid robot. In order to reduce the noise and minimize the risk of falling, we proposed a binary search based spline fitting algorithm for smoothing the joint trajectories which also takes care of collision avoidance with the surroundings. Unlike other sampling-based algorithm, the proposed framework can perform in real-time. 
The framework was tested on a 27 DoF humanoid with an articulated torso in 4 different environment settings. 
Results show the effectiveness and robustness of this approach in learning coordinated tasks in varied set of environments. Further simulation results show the usefulness of the articulated torso in such tasks.

Present work focusses on motion planning in the presence of static obstacles. In our future work, we plan on extending the proposed framework for dynamic obstacles. Also, we plan to learn coordinated tasks which involves movements of the legs as well. 
\bibliographystyle{IEEEtran}
\bibliography{root}
\end{document}